\pdfoutput=1
\documentclass[11pt]{article}

\usepackage[]{EMNLP2022}

\usepackage{times}
\usepackage{latexsym}

\usepackage[T1]{fontenc}
\usepackage[utf8]{inputenc}

\usepackage{microtype}

\usepackage{bbding}
\usepackage{multirow}
\usepackage{booktabs} 
\usepackage{graphicx}
\usepackage{xcolor,colortbl, soul}
\usepackage{arydshln}

\definecolor{berry}{HTML}{e897a8}
\definecolor{apricot}{HTML}{ec9d72}
\definecolor{raspberry}{HTML}{BD6DA2}

\DeclareRobustCommand{\hlone}[1]{{\sethlcolor{apricot}\hl{#1}}}
\DeclareRobustCommand{\hltwo}[1]{{\sethlcolor{raspberry}\hl{#1}}}

\title{Analyzing the Mono- and Cross-Lingual Pretraining Dynamics\\ of Multilingual Language Models}

\author{Terra Blevins$^{1}$ \quad Hila Gonen$^{1,2}$ \quad Luke Zettlemoyer$^{1,2}$ \\
        $^{1}$ Paul G. Allen School of Computer Science \& Engineering, University of Washington \\
        $^{2}$ Meta AI Research \\
        {\tt \{blvns, lsz\}@cs.washington.edu}\\
        {\tt hilagnn@gmail.com}}

\begin{document}
\maketitle
\begin{abstract}
The emergent cross-lingual transfer seen in multilingual pretrained models has sparked significant interest in studying their behavior. However, because these analyses have focused on fully trained multilingual models, little is known about the dynamics of the multilingual pretraining process. We investigate \textit{when} these models acquire their in-language and cross-lingual abilities by probing checkpoints taken from throughout XLM-R pretraining, using a suite of linguistic tasks. Our analysis shows that the model achieves high in-language performance early on, with lower-level linguistic skills acquired before more complex ones. In contrast, the point in pretraining when the model learns to transfer cross-lingually differs across language pairs. Interestingly, we also observe that, across many languages and tasks, the final model layer exhibits significant performance degradation over time, while linguistic knowledge propagates to lower layers of the network. Taken together, these insights highlight the complexity of multilingual pretraining and the resulting varied behavior for different languages over time.
\end{abstract}

\begin{table*}[t]
    \centering
    \small
    \aboverulesep = 0pt
    \belowrulesep = 0pt

    \begin{tabular}{c| c c c | >{\centering\arraybackslash}m{1.5in}}
    \toprule
    \multirow{2}{*}{\textbf{Task}} & \multirow{2}{*}{\textbf{Setup}} & \multicolumn{2}{c |}{\textbf{Num. Langs (Pairs)}} & \multirow{2}{*}{\textbf{Example}} \\
    & & \textbf{In-lang.} & \textbf{X-lang.} & \\
    \hline
    BPC & Masked LM & 94 & -- & \includegraphics[scale=0.5]{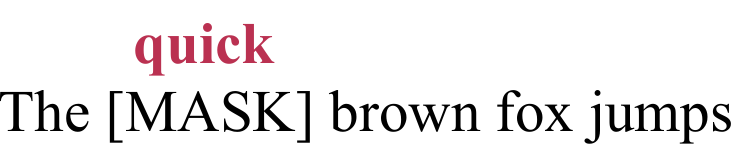} \\
    \hdashline
    POS Tagging & Token Labeling & 44 & 18 $\rightarrow$ 18 & \includegraphics[scale=0.5]{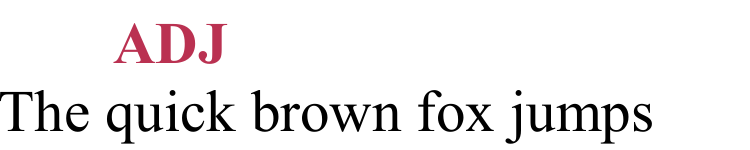} \\
    \hdashline
   Dependency Arc Pred. & Token Pair Labeling & 44 & 18 $\rightarrow$ 18 & \includegraphics[scale=0.5]{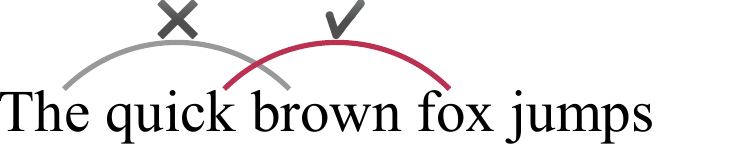} \\
   \hdashline
    Dependency Arc Class. & Token Pair Labeling & 44 & 18 $\rightarrow$ 18 & \includegraphics[scale=0.5]{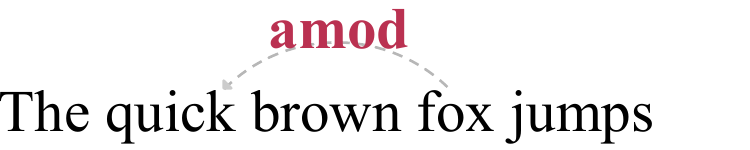} \\
    \hdashline
    XNLI & Sent. Pair Labeling & 15 & 15 $\rightarrow$ 15 & \includegraphics[scale=0.5]{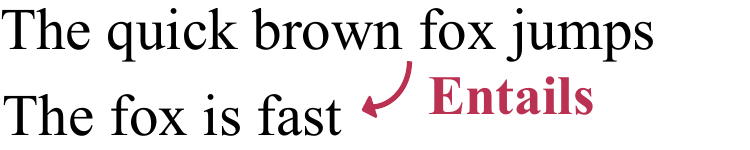} \\
    \hdashline
    SimAlign & Unsupervised Alignment & -- & 1 $\rightarrow$ 6 & \includegraphics[scale=0.5]{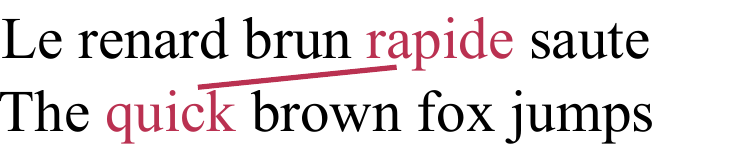} \\
    \toprule
    \end{tabular}
    \caption{Summary of the linguistic information we probe XLM-R$_{replica}$ for throughout pretraining.}
    \label{tab:task-summary}
\end{table*}

\section{Introduction}

Large-scale language models pretrained jointly on text from many different languages \cite{delvin2019mbert, lample2019cross, lin2021few} perform very well on various languages and on cross-lingual transfer between them \cite[e.g.,][]{kondratyuk2019languages, pasini2021xl}. 
Due to this success, there has been a great deal of interest in uncovering what these models learn from the multilingual pretraining signal (\S \ref{sec:background}). However, these works analyze a single model artifact: the final training checkpoint at which the model is considered to be converged. Recent work has also studied monolingual models by expanding the analysis to multiple pretraining checkpoints to see how model knowledge changes across time \cite{liu2021probing}. 

We analyze multilingual training checkpoints throughout the pretraining process in order to identify when multilingual models obtain their in-language and cross-lingual abilities. The case of multilingual language models is particularly interesting, as the model learns both to capture individual languages and to transfer between them just from unbalanced multitask language modeling for each language.

Specifically, we retrain a popular multilingual model, XLM-R \cite{conneau2020unsupervised}, and run a suite of linguistic tasks covering 59 languages on checkpoints from across the pretraining process.\footnote{The XLM-R$_{replica}$ checkpoints are available at \url{https://nlp.cs.washington.edu/xlmr-across-time}.}
This suite evaluates different syntactic and semantic skills in both monolingual and cross-lingual transfer settings. While our analysis primarily focuses on the knowledge captured in model output representations over time, we also consider how the performance of internal layers changes during pretraining for a subset of tasks.

Our analysis uncovers several insights into multilingual knowledge acquisition. First, while the model acquires most in-language linguistic information early on, cross-lingual transfer is learned across the entire pretraining process. Second, the order in which the model acquires linguistic information for each language is generally consistent with monolingual models: lower-level syntax is learned prior to higher-level syntax and then semantics. In comparison, the order in which the model learns to transfer linguistic information between specific languages can vary wildly.

Finally, we observe significant degradation of performance for many languages at the final layer of the last, converged model checkpoint.
However, lower layers of the network often continue to improve later in pretraining and outperform the final layer, particularly for cross-lingual transfer. These observations indicate that there is not a single time step (or layer) in pretraining that performs the best across all languages and suggest that methods that better balance these tradeoffs could improve multilingual pretraining in the future.

\section{Analyzing Knowledge Acquisition Throughout Multilingual  Pretraining}

Our goal is to quantify when information is learned by multilingual models across pretraining. To this end, we reproduce a popular multilingual pretrained model, XLM-R -- referred to as XLM-R$_{replica}$ -- and retain several training checkpoints (\S \ref{sec:replica}). A suite of linguistic tasks is then run on the various checkpoints (\S \ref{sec:li-tasks}).
For a subset of these tasks, we also evaluate at which layer in the network information is captured during pretraining.

Since we want to identify what knowledge is gleaned from the pretraining signal, each task is evaluated without finetuning. The majority of our tasks are tested via \textit{probes}, in which representations are taken from the final layer of the frozen checkpoint and used as input features to a linear model trained on the task of interest \cite{belinkov2020interpretability}.
Additional evaluations we consider for the model include an intrinsic evaluation of model learning (BPC) and unsupervised word alignment of model representations. Each of the tasks in our evaluation suite tests the extent to which a training checkpoint captures some form of \textit{linguistic information}, or a specific aspect of linguistic knowledge, and they serve as a proxy for language understanding in the model.

\subsection{Replicating XLM-R}
\label{sec:replica}
Analyzing model learning throughout pretraining requires access to intermediate training checkpoints, rather than just the final artifact. We replicate the base version of XLM-R and save a number of checkpoints throughout the training process. Our pretraining setup primarily follows that of the original XLM-R, with the exception that we use a smaller batch size (1024 examples per batch instead of 8192) due to computational constraints. All other hyperparameters remain unchanged.

XLM-R$_{replica}$ is also trained on the same data as the original model, CC100. This dataset consists of filtered Common Crawl data for 100 languages, with a wide range of data quantities ranging from 0.1 GiB for languages like Xhosa and Scottish Gaelic to over 300 Gib for English.
As with XLM-R, we train on CC100 for 1.5M updates and save 39 checkpoints for our analysis, with more frequent checkpoints taken in the earlier portion of training: we save the model every 5k training steps up to the 50k step, and then every 50k steps. Further details about the data and pretraining scheme can be found in \citet{conneau2020unsupervised}. We compare the final checkpoint of XLM-R$_{replica}$ to the original XLM-R$_{base}$ and find that while XLM-R$_{replica}$ performs slightly worse in-language, the two models perform similarly cross-lingually (Appendix \ref{app:exp-details}).

\begin{figure*}
    \centering
    \includegraphics[width=\textwidth]{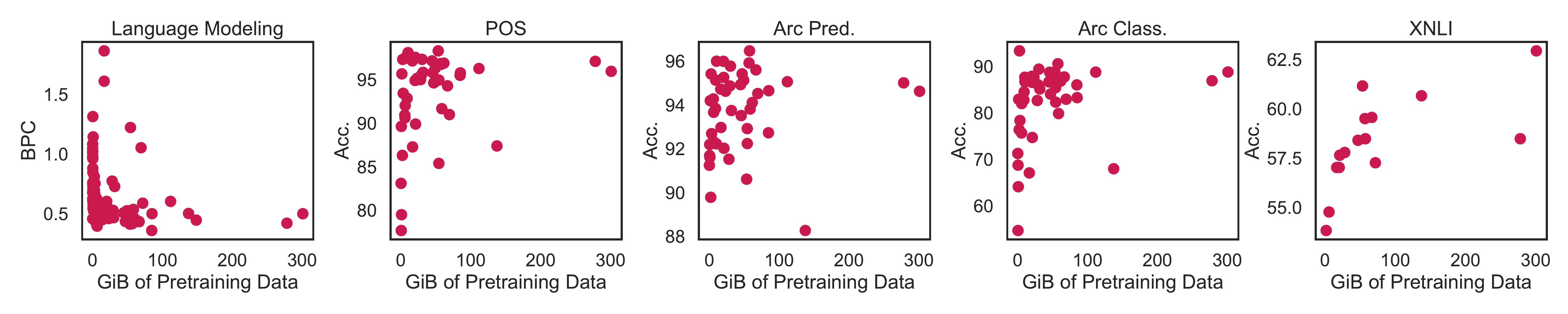}
    \caption{Best in-language performance of XLM-R$_{replica}$ on various tasks and languages across all checkpoints.}
    \label{fig:best_performance}
\end{figure*}

\subsection{Linguistic Information Tasks}
\label{sec:li-tasks}
The analysis suite covers different types of syntactic knowledge, semantics in the form of natural language inference, and word alignment (Table \ref{tab:task-summary}).
These tasks evaluate both in-language linguistics as well as cross-lingual transfer with a wide variety of languages and language pairs. Unless otherwise stated (\S \ref{sec:layerwise}), we evaluate the output from the final layer of XLM-R$_{replica}$. Additionally, most tasks (POS tagging, dependency structure tasks, and XNLI) are evaluated with accuracy; the MLM evaluation is scored on BPC, and SimAlign is evaluated on F1 performance.
Appendix \ref{app:exp-details} details the languages covered by each of these tasks and further experimental details.

\paragraph{MLM Bits per Character (BPC)}
As an intrinsic measure of model performance, we consider the bits per character (BPC) on each training language of the underlying MLM. For a sequence \textbf{s}, BPC(\textbf{s}) is the (average) negative log-likelihood (NLL) of the sequence under the model normalized by the number of characters per token; lower is better for this metric.  These numbers are often not reported for individual languages or across time for multilingual models, making it unclear how well the model captures each language on the pretraining task. We evaluate BPC on the validation split of CC100. 

\paragraph{Part-of-Speech (POS) Tagging} 
We probe XLM-R$_{replica}$ with a linear model mapping the representation for each word to its corresponding POS tag; words that are split into multiple subword tokens in the input are represented by the average of their subword representations. The probes are trained using the Universal Dependencies (UD) treebanks for each language \cite{nivre2020universal}. For cross-lingual transfer, we evaluate a subset of languages that occur in Parallel Universal Dependencies (PUD; \citealp{zeman2017conll}), a set of parallel test treebanks, to control for any differences in the evaluation data. 

\paragraph{Dependency Structure}
We evaluate syntactic dependency structure knowledge with two pair-wise probing tasks: \textit{arc prediction}, in which the probe is trained to identify pairs of words that are linked with a dependency arc; and \textit{arc classification}, where the probe labels a pair of words with their corresponding dependency relation.
The two word-level representations $r_1$ and $r_2$ are formatted as a single concatenated input vector $[r_1; r_2; r_1 \odot r_2]$, following \citet{blevins2018deep}. 
This combined representation is then used as the input to a linear model that labels the word pair.
Probes for both dependency tasks are trained and evaluated with the same set of UD treebanks as POS tagging.

\paragraph{XNLI} 
We also consider model knowledge of natural language inference (NLI), where the probe is trained to determine whether a pair of sentences entail, contradict, or are unrelated to each other. Given two sentences, we obtain their respective representation $r_1$ and $r_2$ by averaging all representations in the sentence, and train the probe on the concatenated representation $[r_1; r_2; r_1 \odot r_2]$. We train and evaluate the probes with the XNLI dataset \cite{conneau2018xnli}; for training data outside of English, we use the translated data provided by \citet{singh2019bert}.

\paragraph{Word Alignment} 
In the layer-wise evaluation (\S \ref{sec:layerwise}), we evaluate how well the model's internal representations are aligned using SimAlign \cite{sabet2020simalign}, an unsupervised algorithm for aligning bitext at the word level using multilingual representations. We evaluate the XLM-R$_{replica}$ training checkpoints with SimAlign on manually annotated reference alignments for the following language pairs: EN-CS \cite{marecek2008automatic}, EN-DE\footnote{Gold alignments on EuroParl \cite{koehn2005europarl}, http://www-i6.informatik.rwth-aachen.de/goldAlignment/}, EN-FA \cite{tavakoli2014phrase}, EN-FR \cite[WPT2003,][]{och2000improved}, EN-HI\footnote{\label{wpt2005} WPT2005, http://web.eecs.umich.edu/~mihalcea/wpt05/}, and EN-RO$^{\ref{wpt2005}}$. 

\section{In-language Learning Throughout Pretraining}

We first consider the in-language, or monolingual, performance of  XLM-R$_{replica}$ on different types of linguistic information across pretraining.
We find that in-language linguistics is learned (very) early in pretraining and is acquired in a consistent order, with lower-level syntactic information learned before more complex syntax and semantics. Additionally, the final checkpoint of XLM-R$_{replica}$ often experiences performance degradation compared to the best checkpoint for a language, suggesting that the model is forgetting information for a number of languages by the end of pretraining.

\subsection{Monolingual Performance for Different Languages}
\label{sec:in-lang-performance}
Figure \ref{fig:best_performance} presents the overall best performance of the model across time on the considered tasks and languages. We observe a large amount of variance in performance  on each task. Across languages, XLM-R$_{replica}$ performance ranges between 1.86 and 0.36 BPC for language modeling, 88.3\% and 96.5\% accuracy for dependency arc prediction, 77.67\% and 98.3\% accuracy for POS tagging, 54.7\% and 93.3\% accuracy for arc classification, and 53.8\% and 62.9\% accuracy for XNLI.
Overall, these results confirm previous findings that multilingual model performance varies greatly on different languages (\S \ref{sec:background}). 

\begin{figure}
    \centering
    \includegraphics[width=0.6\linewidth]{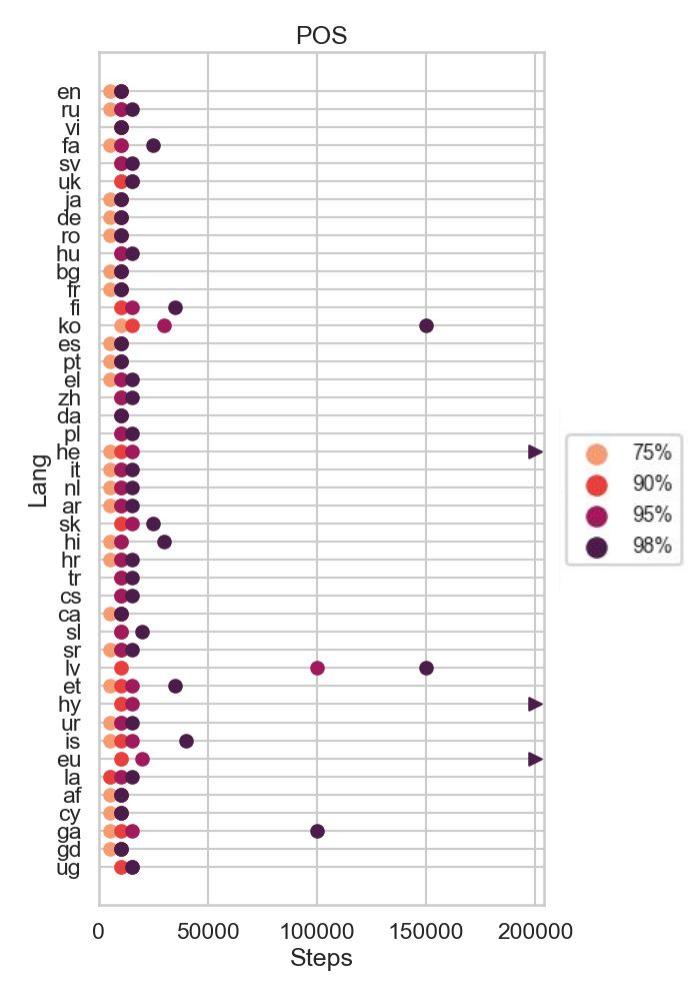}
    \caption{Learning progress of XLM-R$_{replica}$ on POS tagging, up to 200k training steps. Each point represents the step at which the model achieves x\% of the best overall performance of the model on that task; arrows indicate languages that reach the 98\% mark after 200k steps.}
    \label{fig:progress_scatter}
\end{figure}

\subsection{When Does XLM-R Learn Linguistic Information?}
\label{sec:when-in-lang}
Figure \ref{fig:progress_scatter} shows the step at which XLM-R$_{replica}$ reaches different percentages of its best performance of the model on POS tagging. Figures for the other tasks are given in Appendix \ref{app:more-analysis}.

\paragraph{Monolingual linguistics is acquired early in pretraining}
We find that XLM-R$_{replica}$ acquires the majority of in-language linguistic information early in training. However, the average time step for acquisition varies across tasks. For dependency arc prediction, all languages achieve 98\% or more of total performance by 20k training steps (out of 1.5M total updates). In contrast, XNLI is learned later with the majority of the languages achieving 98\% of the overall performance after 100k training updates. This order of acquisition is in line with monolingual English models, which have also been found to learn syntactic information before higher-level semantics  \cite{liu2021probing}.

We also observe that this order of acquisition is often maintained within individual languages. 12 out of 13 of the languages shared across all tasks reach 98\% of the best performance consistently in the order of POS tagging and arc prediction (which are typically learned within one checkpoint of each other), arc classification, and XNLI. 

\begin{figure}
    \centering
    \includegraphics[width=0.85\linewidth]{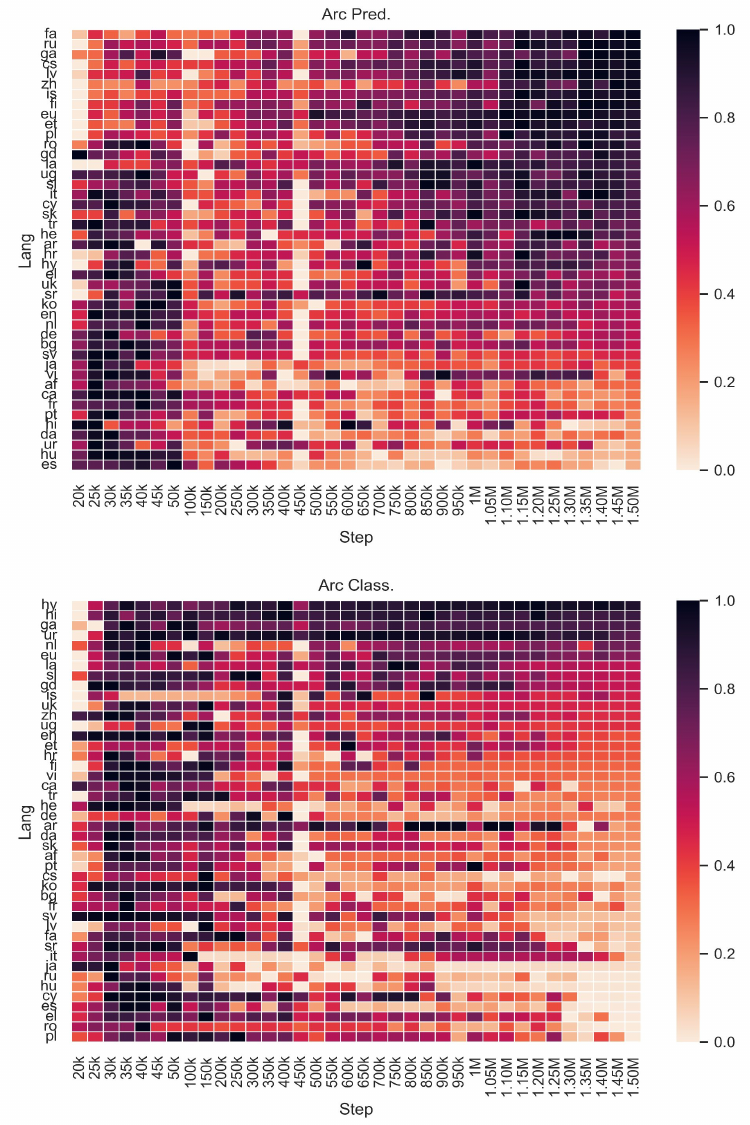}
    \caption{Heatmap of relative performance over time for dependency arc prediction and classification. Languages are ordered by performance degradation in the final training checkpoint.}
    \label{fig:progress_heatmap}
\end{figure}

\paragraph{Model behavior later in pretraining varies across languages}
For some languages and tasks, XLM-R$_{replica}$ never achieves good absolute performance (Figure \ref{fig:best_performance}). For others, the performance of XLM-R$_{replica}$ decreases later in pretraining, leading the converged model to have degraded performance on those tasks and languages (Figure \ref{fig:progress_heatmap}).

We hypothesize that this is another aspect of the ``curse of multilinguality,'' where some languages are more poorly captured in multilingual models due to limited model capacity \cite{conneau2020unsupervised, wang2020negative}, that arises during the training process. We also find that the ranking of languages by performance degradation is not correlated across tasks.
This suggests the phenomenon is not limited to a subset of low-resource languages and can affect any language learned by the model.

More generally, these trends demonstrate that the best model state varies across languages and tasks. 
Since BPC continues to improve on all individual training languages throughout pretraining (Appendix \ref{app:more-analysis}), the results also indicate that performance on the pretraining task is not directly tied to performance on the linguistic probes. This is somewhat surprising, given the general assumption that better pretraining task performance corresponds to better downstream task performance. 

\begin{figure*}
    \centering
    \includegraphics[width=\textwidth]{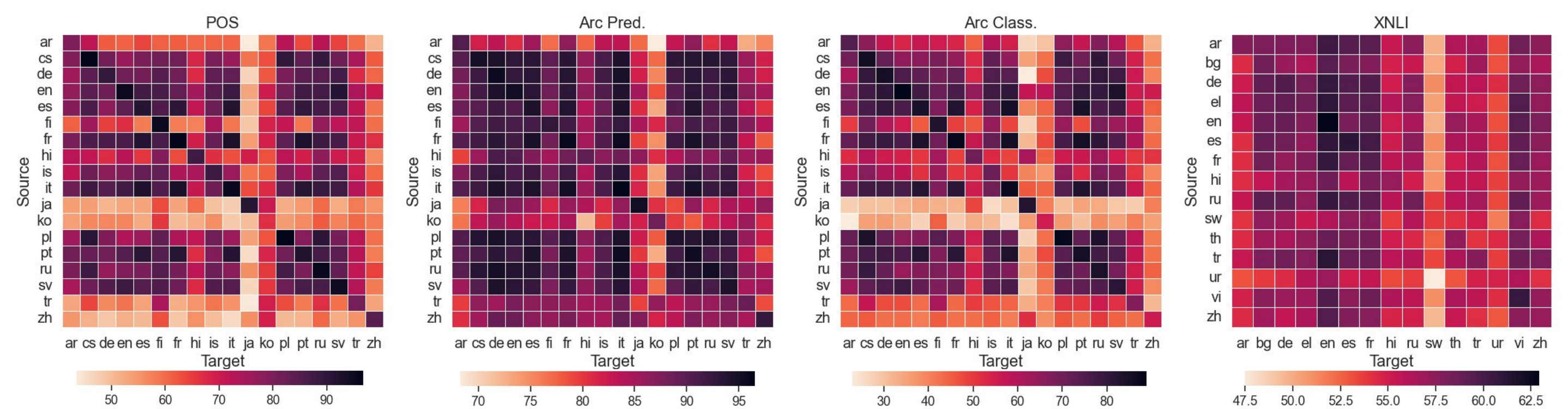}
    \caption{Overall performance of XLM-R$_{replica}$ on each analysis task when transferring from various source to target languages.}
    \label{fig:xl_heatmap}
\end{figure*}

\begin{figure*}
    \centering
    \includegraphics[width=\textwidth]{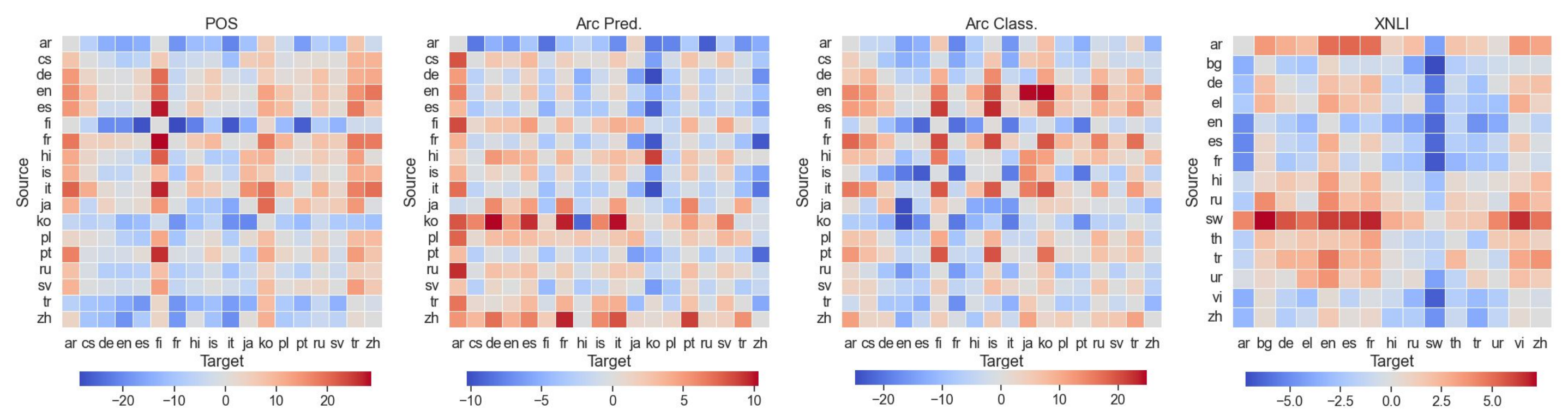}
    \caption{Heatmap of the asymmetry of cross-lingual transfer in XLM-R$_{replica}$. Each cell shows the difference in performance between language pairs ($l_1 \rightarrow l_2$) and ($l_2 \rightarrow l_1$). }
    \label{fig:asym_heatmap}
\end{figure*}

\begin{figure*}
    \centering
    \includegraphics[width=\textwidth]{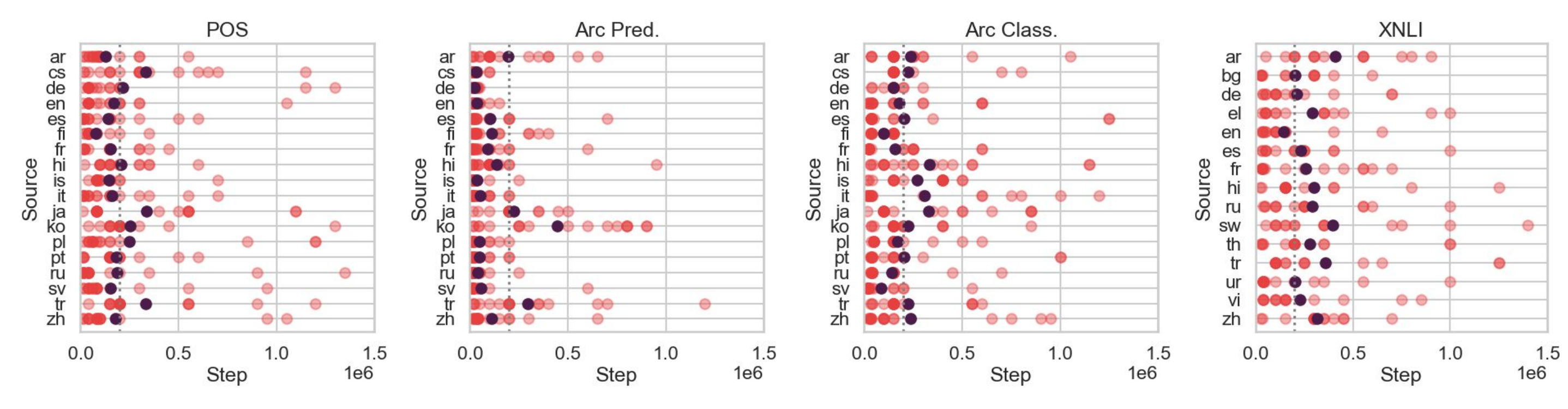}
    \caption{Cross-lingual learning progress of XLM-R$_{replica}$ across pretraining. Each red point represents the step to 98\% of the best performance for a language pair; the purple represents the mean 98\% transfer step for the source language.}
    \label{fig:xl_progress_scatter}
\end{figure*}

\begin{figure*}
    \centering
    \includegraphics[width=\textwidth]{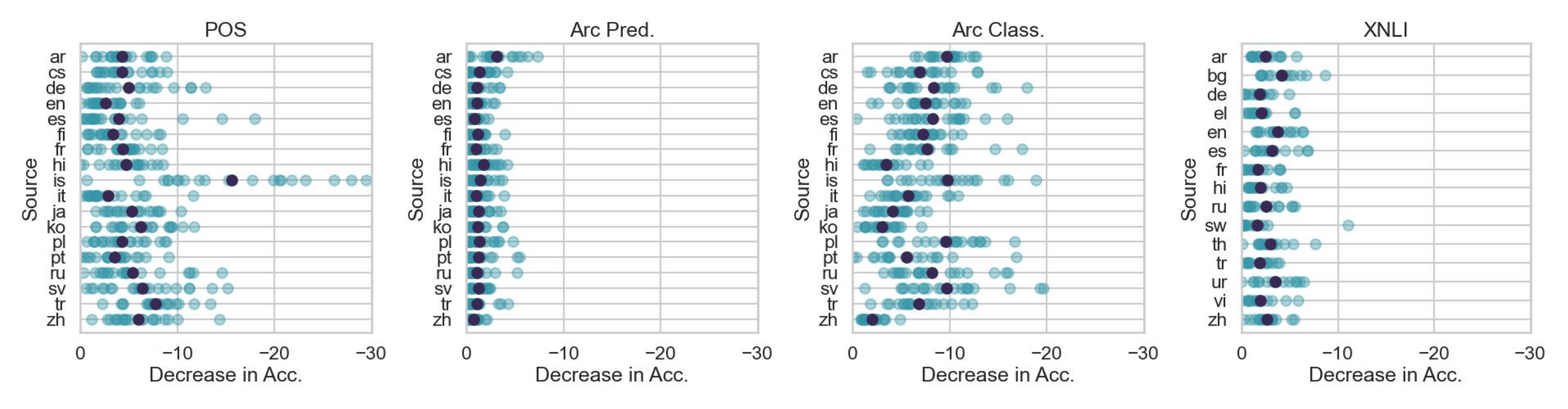}
    \caption{Degradation of cross-lingual transfer performance of XLM-R$_{replica}$ across pretraining. Each blue point represents the change in performance from the overall best step to the final model checkpoint for a language pair; the navy represents the mean decrease for the source language.}
    \label{fig:xl_forgetting_scatter}
\end{figure*}

\section{Cross-lingual Transfer Throughout Pretraining}
\label{sec:xl-transfer}

Another question of interest is: when do multilingual models learn to transfer between languages?
We find that cross-lingual transfer is acquired later in pretraining than monolingual linguistics and that the step at which XLM-R$_{replica}$ learns to transfer a specific language pair varies greatly. Furthermore, though the order in which XLM-R$_{replica}$ learns to transfer different linguistic information across languages is on average consistent with in-language results, the order in which the model learns to transfer across specific language pairs for different tasks is much more inconsistent.

\subsection{Overall Transfer Across Language Pairs}
\label{sec:overall-transfer}
\paragraph{Which languages transfer well?} Figure \ref{fig:xl_heatmap} shows cross-lingual transfer between different language pairs; most source languages perform well in-language (the diagonal).  We observe that some tasks, specifically dependency arc prediction, are easier to transfer between languages than others; however, across the three tasks with shared language pairs (POS tagging, arc prediction, and arc classification) we see similar behavior in the extent to which each language transfers to others. For example, English and Italian both transfer well to most of the target languages. However, other languages are isolated and do not transfer well into or out of other languages, even though in some cases, such as Japanese, the model achieves good in-language performance. 

On XNLI there is more variation in in-language performance than is observed on the syntactic tasks. This stems from a more general trend that some languages appear to be easier to transfer into than others, leading to the observed performance consistency within columns. For example, English appears to be particularly easy for XLM-R$_{replica}$ to transfer into, with 12 out of the 14 non-English source languages performing as well or better on English as in-language. 

\paragraph{Cross-lingual transfer is asymmetric} We also find that language transfer is asymmetric within language pairs (Figure \ref{fig:asym_heatmap}). There are different transfer patterns between dependency arc prediction and the other syntactic tasks: for example, we see that Korean is worse relatively as a source language than as the target for POS tagging and arc classification, but performs better when transferring to other languages in arc prediction. However, other languages such as Arabic have similar trends across the syntactic tasks.
On XNLI, we find that Swahili and Arabic are the most difficult languages to transfer into, though they transfer to other languages reasonably well. 

These results expand on observations in \citet{turc2021revisiting} and emphasize that the choice of source language has a large effect on cross-lingual performance in the target. However, there are factors in play in addition to linguistic similarity causing this behavior, leading to asymmetric transfer within a language pair. We further examine these correlations  with overall cross-lingual performance and asymmetric transfer in \S \ref{app:cross-lang-corr}.

\subsection{When is Cross-lingual Transfer Learned During Pretraining?}
\label{sec:when-transfer}

We next consider when during pretraining XLM-R$_{replica}$ learns to transfer between languages (Figure \ref{fig:xl_progress_scatter}; the dotted line indicates the 200k step cutoff used in Figure \ref{fig:progress_scatter} for comparison). Unlike the case of monolingual performance, the step at which the model acquires most cross-lingual signal (98\%) varies greatly across language pairs. We also find that (similar to the in-language setting) higher-level linguistics transfer later in pretraining than lower-level ones: the average step for a language pair to achieve 98\% of overall performance occurs at 115k for dependency arc prediction, 200k for POS tagging, 209k for dependency arc classification, and 274k for XNLI. In contrast, when the model learns to transfer different linguistic information between two specific languages can vary wildly: only approximately 21\% of the language pairs shared across the four tasks transfer in the expected order. 

We also investigate the amount to which the cross-lingual abilities of XLM-R$_{replica}$ decrease over time (Figure \ref{fig:xl_forgetting_scatter};  more detailed across time results for transferring out of English are given in Appendix \ref{app:more-analysis}).  
Similarly to in-language behavior, we find that the model exhibits notable performance degradation for some language pairs (in particular on POS tagging and dependency arc classification), and the extent of forgetting can vary wildly across target languages for a given source language. 

\begin{figure}
    \centering
    \includegraphics[width=\linewidth]{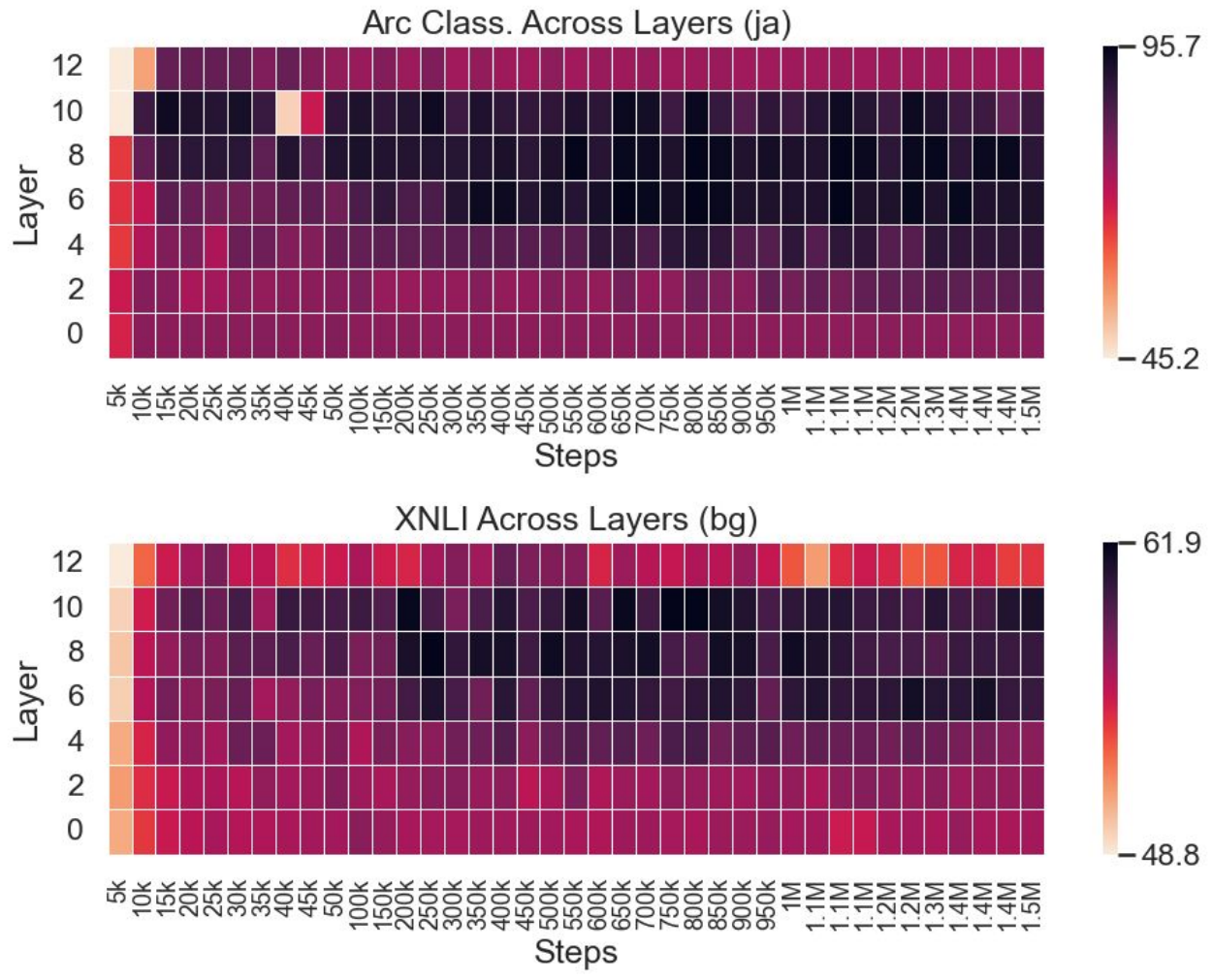}
    \caption{Heatmap of XLM-R$_{replica}$ performance for Japanese arc classification and Bulgarian XNLI. Additional heatmaps are given in Appendix \ref{app:layerwise}.}
    \label{fig:in-lang-layer}
\end{figure}

\section{Layer-wise Learning Throughout Pretraining}
\label{sec:layerwise}
In the experiments above we show that in many cases the final layer of  XLM-R$_{replica}$ forgets information by the end of pretraining. Motivated by this, we investigate whether this information is retained in a different part of the network by probing how information changes \textit{across} layers during pretraining. We find a surprising trend in how the best-performing layer changes over time: the model acquires knowledge in higher layers early on, which then propagates to and improves in the lower layers later in pretraining.

\subsection{In-language Knowledge Across Layers}
\label{sec:in-lang-layerwise}
We first look at the layer-wise performance of XLM-R$_{replica}$ on a subset of languages for dependency arc classification (CS, EN, HI, and JA) and XNLI (BG, EN, HI, and ZH) (Figure \ref{fig:in-lang-layer}). We find that the last layer is often not the best one for each task, with lower layers often outperforming the final one. On average, the best internal layer state outperforms the final layer of XLM-R$_{replica}$ by 7.59 accuracy points on arc classification and 2.93 points on XNLI.

We also observe a trend of lower layers acquiring knowledge later in training than the final one. To investigate this, we calculate the expected best layer (i.e., the average layer weighted by performance) at each checkpoint and find that it decreases over time, by up to 2.79 layers for arc classification and 2.49 layers for XNLI (Appendix \ref{app:layerwise}), indicating that though the final layer quickly fits to the forms of in-language information we test for, this information then shifts to lower layers in the network over time.

\begin{figure}
    \centering
    \includegraphics[width=\linewidth]{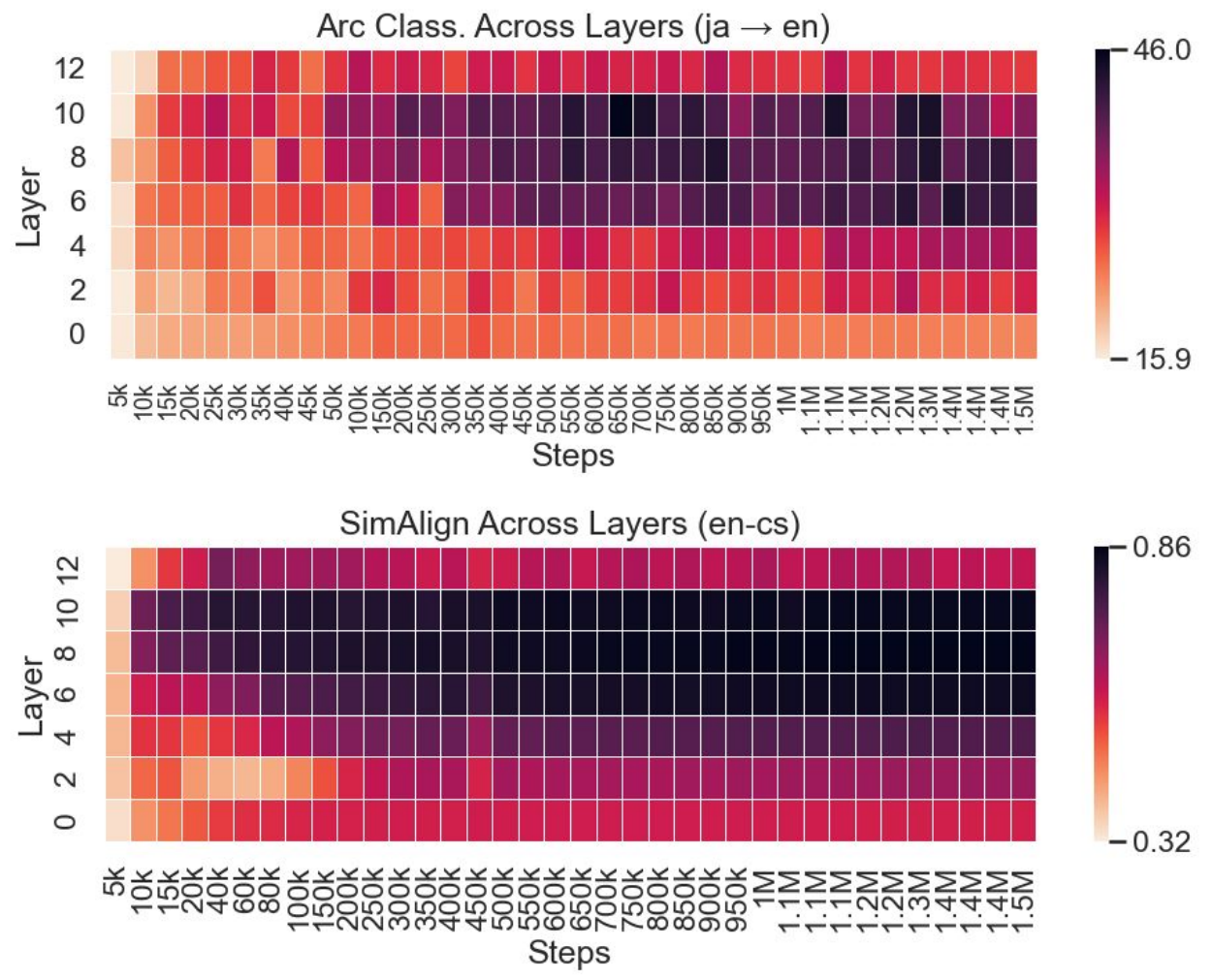}
    \caption{Heatmap of XLM-R$_{replica}$ cross-lingual performance by layer for arc classification (JA → EN) and SimAlign (EN-CS).}
    \label{fig:xl-layerwise-main}
\end{figure}

\subsection{Cross-lingual Knowledge Across Layers}

Next, we consider how cross-lingual transfer skills are captured across layers during pretraining. Every other XLM-R$_{replica}$ layer  is evaluated on the subsets of languages for arc classification and XNLI in \S \ref{sec:in-lang-layerwise}. We also use SimALign to test how well word representations at these layers align from English to \{CS, DE, FA, FR, HI, RO\}.% (\S \ref{sec:li-tasks}).
We observe similar trends with respect to layer performance over time to the in-language results (Figure \ref{fig:xl-layerwise-main}; additional results given in Appendix \ref{app:layerwise}). Specifically, we observe an average decrease in the expected layer of 1.10 (ranging from 0.67 to 2.20) on arc classification, 1.02 (ranging from 0.37 to 2.01) on XNLI, and 1.66 (ranging from 0.83 to 2.41) on SimAlign. 

We also observe that while most layers perform relatively well  in-language performance, the lowest layers of XLM-R$_{replica}$ (layers 0-4) often perform much worse than the middle and final layers for cross-lingual transfer throughout the pretraining process -- for example, in the case of Japanese to English on arc classification. 
We hypothesize that this is due to better alignment across languages in later layers, similar to the findings in \citet{muller2021first}.

\section{Related Work}
\label{sec:background}

\paragraph{Linguistic knowledge in multilingual models}
There have been several different approaches to quantifying the linguistic information that is learned by multilingual models. One direction has performed layer-wise analyses to quantify what information is stored at different layers in the model \cite{de2020special, taktasheva2021shaking, papadimitriou2021deep}. Others have examined the extent to which the different training languages are captured by the model, finding that some languages suffer in the multilingual setting despite the overall good performance exhibited by the models \cite{conneau2020unsupervised, wang2020negative}.

\paragraph{Cross-lingual transfer in multilingual models}
Another line of analysis seeks to understand the cross-lingual abilities of multilingual models. \citet{chi2020finding} show that subspaces of mBERT representations that capture syntax are approximately shared across languages, suggesting that portions of the model are cross-lingually aligned.
A similar direction of interest is whether multilingual models learn language-agnostic representations. \citet{singh2019bert} find that mBERT representations can be partitioned by language, indicating that the representations retain language-specific information. Similarly, other work has shown that mBERT representations can be split into language-specific and language-neutral components \cite{libovicky2019language, gonen2020greek, muller2021first}. 

Other work has investigated the factors that affect cross-lingual transfer.
These factors include the effect of sharing subword tokens on cross-lingual transfer \cite{conneau2020emerging, k2020cross, deshpande2021bert} and which languages act as good source languages for cross-lingual transfer \cite{turc2021revisiting}. Notably, \citet{lauscher2020zero}, \citet{k2020cross} and \citet{hu2020xtreme} find that multilingual pretrained models perform worse when transferring to distant languages and low-resource languages.

\paragraph{Examining Pretrained Models Across Time}
A recent direction of research has focused on probing multiple checkpoints taken from different points in the pretraining process, in order to quantify when the model learns information. These works have examined the acquisition of syntax \cite{perez2021much} as well as higher-level semantics and world knowledge over time \cite{liu2021probing} from the RoBERTa pretraining process. Similarly, \citet{chiang2020pretrained} perform a similar temporal analysis for AlBERT, and \citet{choshen2022grammar} find that the order of linguistic acquisition during language model training is consistent across model sizes, random seeds, and LM objectives.

Most work on probing pretrained models across the training process has focused on monolingual, English models. There are some limited exceptions: \citet{dufter2020identifying} present results for multilingual learning in a synthetic bilingual setting, and \citet{wu2020all} examine performance across pretraining epochs for a small number of languages. 
However, this paper is the first to report a comprehensive analysis of monolingual and cross-lingual knowledge acquisition on a large-scale multilingual model.

\section{Conclusion}

In this paper, we probe training checkpoints across time to analyze the training dynamics of the \mbox{XLM-R} pretraining process. 
We find that although the model learns in-language linguistic information early in training  -- similar to findings on monolingual models -- cross-lingual transfer is obtained all throughout the pretraining process. 

Furthermore, the order in which linguistic information is acquired by the model is generally consistent, with lower-level syntax acquired before semantics. However, we observe that for individual language pairs this order can vary wildly, and our statistical analyses demonstrate that model learning speed and overall performance on specific languages (and pairs) are difficult to predict from language-specific factors.

We also observe that the final model artifact of XLM-R$_{replica}$ performs often significantly worse than earlier training checkpoints on many languages and tasks. However, layer-wise analysis of the model shows that linguistic information shifts lower in the network during pretraining, with lower layers eventually outperforming the final layer. 
Altogether, these findings provide a better understanding of multilingual training dynamics that can inform future pretraining approaches.

\section{Limitations}
We note some potential limitations of this work. We consider a single pretraining setting (replicating the training of XLM-R$_{base}$), and the extent to which our findings transfer to other multilingual pretraining settings remains an open question. In particular, pretraining a language model with more parameters or on different multilingual data could lead to other trends, though many of our findings are consistent with prior work. 

Additionally, despite our attempts to use diverse datasets for evaluating these models, the language choices available in annotated NLP data are skewed heavily towards Indo-European, especially English and other Western European, languages. This means that many of the low-resource languages seen in the pretraining data are unaccounted for in this study. Due to this, we only evaluate word alignment between six languages paired with English, and a number of the non-English datasets we use are translated from English.

Another product of limited multilingual resources is our ability to compare across languages; in UD, each treebank is annotated on different domains with different dataset sizes. This limits the comparisons we can make across probe training settings, though we focus on changes within individual languages in this work. To address this limitation, we use the parallel test sets from Parallel Universal Dependencies for our cross-lingual transfer experiments, which allows us to compare performance on different target languages from the same source language directly.

\section*{Acknowledgements}
We would like to thank Naman Goyal for his advice on retraining XLM-R and Masoud Jalili Sabet for providing the data to run SimAlign. We also thank Leo Z. Liu for helpful conversations on this work as well as the anonymous reviewers for their thoughtful feedback.

\bibliography{anthology,custom}
\bibliographystyle{acl_natbib}

\appendix

\begin{table}[b!]
    \centering
    \small
    \aboverulesep = 0pt
    \belowrulesep = 0pt
    \begin{tabular}{p{0.12\linewidth} | p{0.78\linewidth}}
        \toprule
       \textbf{Task} & \textbf{Languages} \\
        \hline
        \textbf{BPC} & af, am, ar, as, az, be, bg, bn, br, bs, ca, cs, cy, da, de, el, en, eo, es, et, eu, fa, fi, fr, fy, ga, gd, gl, gu, ha, he, hi, hr, hu, hy, id, is, it, ja, jv, ka, kk, km, kn, ko, ku, ky, la, lo, lt, lv, mg, mk, ml, mn, mr, ms, my, ne, nl, no, om, or, pa, pl, ps, pt, ro, ru, sa, sd, si, sk, sl, so, sq, sr, su, sv, sw, ta, te, th, tl, tr, ug, uk, ur, uz, vi, xh, yi, zh, zh \\
        \hline
        \textbf{UD} & af, \textbf{ar}, bg, ca, \textbf{cs}, cy, da, \textbf{de}, el, \textbf{en}, \textbf{es}, et, eu, fa, \textbf{fi}, \textbf{fr}, ga, gd, he, \textbf{hi}, hr, hu, hy, \textbf{is}, \textbf{it}, \textbf{ja}, \textbf{ko}, la, lv, nl, \textbf{pl}, \textbf{pt}, ro, \textbf{ru}, sk, sl, sr, \textbf{sv}, \textbf{tr}, ug, uk, ur, vi, \textbf{zh} \\
        \hline
        \textbf{XNLI} & \textbf{ar}, \textbf{bg}, \textbf{de}, \textbf{el}, \textbf{en}, \textbf{es}, \textbf{fr}, \textbf{hi},  \textbf{ru}, \textbf{sw}, \textbf{th},  \textbf{tr}, \textbf{ur}, \textbf{vi},  \textbf{zh} \\
        %\textbf{SimAlign} & en $\rightarrow$ {cs, de, fa, fr, hi, ro} \\
        \toprule
    \end{tabular}
    \caption{Table summarizing the languages considered for each task. Languages in bold are also used for the cross-lingual setting of the task. UD covers all of the languages used for POS tagging, dependency arc prediction, and dependency arc classification.}
    \label{tab:task_langs}
\end{table}

\section{Linguistic Probe Details}
\label{app:exp-details}
Table \ref{tab:task_langs} presents the languages that are included in each of the probing tasks. We filter the Romanized versions of languages from the CC100 dataset, leaving us with 94 for evaluation.

\subsection{Experimental Setup}

Each evaluation is run on the frozen parameters of a training checkpoint of XLM-R$_{replica}$. All representations are taken from the final (12th) layer of the encoder, except for the experiments presented in \S \ref{sec:layerwise}, which consider the performance of different layers within the model over time.

For the linguistic information tasks involving probing, each probe consists of a single linear layer, trained with a batch size of 256 for 50 epochs with early stopping performed on the validation set. The probes therefore consist of a limited number of parameters $m*l$, where $m = 768$ is the output dimension of the model and $l$ is the size of the task label set. Following \citet{liu2019linguistic}, the probes are optimized with a learning rate of 1e-3. Each probe is trained on a single Nvidia V100 16GB GPU and takes between <1 minute and 6 minutes to train (depending on dataset size, which varies by language and task). The reported results for each probe are the averaged performance across five runs.

For SimAlign, we use the default settings provided in the SimAlign implementation.\footnote{https://github.com/cisnlp/simalign} We report word-level alignment performance (instead of sub-word alignment) using the itermax alignment algorithm.

\subsection{XLM-R Replication Details}
XLM-R$_{replica}$ consists of the same model architecture as XLM-R$_{base}$, with a total of 270M parameters. We train the model for 1.5 million updates on 64 Nvidia V100 32 GB GPUs using the fairseq library \cite{ott2019fairseq}. Notably, the language sampling alpha for up-weighting less frequent languages is set to $\alpha=0.7$: this matches the value used for the XLM-R, though it was reported as $\alpha=0.3$ in the original paper.

\begin{table}[]
    \centering
    \aboverulesep = 0pt
    \belowrulesep = 0pt
    \begin{tabular}{c c | c c}
        \toprule
        \multicolumn{2}{c}{\textbf{Task}} & \textbf{XLM-R$_{base}$} & \textbf{XLM-R$_{replica}$} \\
        \hline
        \multirow{3}{*}{In-lang}& BPC & 0.609* & 0.652 \\
        & POS & 89.65* & 87.20 \\
        & XNLI & 58.08$^*$ & 55.73 \\
        \hline
        \multirow{2}{*}{X-lang} & POS & 66.01* & 64.94 \\
        & XNLI & 53.26 & 53.77$^*$ \\
        \toprule
    \end{tabular}
    \caption{Average performance across languages of XLM-R$_{base}$ and the final checkpoint of XLM-R$_{replica}$.}
    \label{tab:app-model-comparison}
\end{table}

\subsection{Comparison with XLM-R$_{base}$}
We also compare the performance of our retrained XLM-R$_{replica}$ model against the original XLM-R$_{base}$ on a subset of the tasks in our evaluation suite (Table \ref{tab:app-model-comparison}). We find that on average, the original XLM-R model achieves better BPC than the replicated model; this is likely due to the decrease in batch size while retraining the model. The replica model also performs slightly worse than the original on in-language tasks but comparably cross-lingually (and outperforms the original model on cross-lingual XNLI).

\begin{table*}[t]
\centering
    \small
    \aboverulesep = 0pt
    \belowrulesep = 0pt
    \begin{tabular}{c c | l l l l l}
    \toprule
    \multirow{2}{*}{\textbf{Variable}}& \multirow{2}{*}{\textbf{Factors}} & \multicolumn{5}{c}{\textbf{Spearman ($\rho$)}}\\
    & & BPC & POS & Arc Pred. & Arc Class. & XNLI  \\
    \hline
    \multirow{3}{*}{Task Perf.} & Pretraining Data & \cellcolor{raspberry} -0.597** & 0.258 & 0.267 & \cellcolor{apricot} 0.411* & \cellcolor{raspberry} 0.767**  \\
    & Task Data & \cellcolor{raspberry} -0.597** & \cellcolor{apricot} 0.462* & 0.276 & \cellcolor{raspberry} 0.527** & \cellcolor{gray}  \\
    & Lang Sim. & \cellcolor{raspberry} 0.427** & \cellcolor{apricot} -0.315* & -0.170 & \cellcolor{apricot} -0.427* & \cellcolor{raspberry} -0.779**  \\
    \toprule
    \multirow{3}{*}{Steps to 95\%} & Pretraining Data & 0.135 & -0.290 & -0.193 & \cellcolor{apricot} -0.301* & -0.239   \\
    & Task Data & 0.135 & -0.065 & -0.260 & -0.209 & \cellcolor{gray}  \\
    & Lang Sim. & \cellcolor{raspberry} -0.385** & 0.156  & 0.268 & \cellcolor{apricot} 0.325* & 0.316  \\
    \hline 
    \multirow{3}{*}{Forgetting} & Pretraining Data & \cellcolor{gray} & 0.230 & 0.218 & \cellcolor{apricot} 0.437* & \cellcolor{apricot} 0.564*   \\
    & Task Data & \cellcolor{gray} & \cellcolor{apricot} -0.322* & \cellcolor{apricot} -0.338* & -0.015 & \cellcolor{gray} \\
    & Lang Sim. & \cellcolor{gray} & 0.172 & -0.158 & -0.181 & \cellcolor{raspberry} -0.795**  \\
    \toprule
    \end{tabular}
    \caption{Correlation study of different factors against measures of in-language knowledge. \hlone{* p < 0.05}, \hltwo{** p < 0.001}}
    \label{tab:in-lang-corr}
\end{table*}

\begin{table*}[t]
\centering
    \small
    \aboverulesep = 0pt
    \belowrulesep = 0pt
    \begin{tabular}{c c | l l l l}
    \toprule
    \multirow{2}{*}{\textbf{Variable}}& \multirow{2}{*}{\textbf{Factors}} & \multicolumn{4}{c}{\textbf{Spearman ($\rho$)}}\\
    & & POS & Arc Pred. & Arc Class. & XNLI  \\
    \hline
    \multirow{4}{*}{Task Perf.} & Src. Pretraining Data & \cellcolor{apricot} 0.113* & 0.107 & \cellcolor{apricot} 0.117* & \cellcolor{apricot} 0.178*  \\
    & Trg. Pretraining Data & 0.038 & \cellcolor{apricot} 0.144* & 0.015 & \cellcolor{raspberry} 0.625**  \\
    & Task Data & \cellcolor{raspberry} 0.245** & \cellcolor{apricot} 0.124* & \cellcolor{apricot} 0.129* & \cellcolor{gray} \\
    & Lang Sim. & \cellcolor{raspberry} -0.598** & \cellcolor{raspberry} -0.575** & \cellcolor{raspberry} -0.593** & \cellcolor{raspberry} -0.321**  \\
    \hline
    \multirow{3}{*}{Asymmetry} & Src. Pretraining Data & \cellcolor{apricot} 0.116* & -0.045 & \cellcolor{apricot} 0.140* & \cellcolor{apricot} -0.423*  \\
    & Trg. Pretraining Data & \cellcolor{apricot} -0.116* & 0.045 & \cellcolor{apricot} -0.140* & \cellcolor{apricot} 0.423* \\
    & Task Data & \cellcolor{apricot} 0.123* & -0.016 & -0.077 & \cellcolor{gray} \\
    \toprule
    \multirow{4}{*}{Steps to 95\%} & Src. Pretraining Data & \cellcolor{raspberry} -0.290** & -0.023 & \cellcolor{apricot} -0.132* & \cellcolor{apricot} -0.195* \\
    & Trg. Pretraining Data & \cellcolor{apricot} -0.123* & -0.066 & -0.106 & -0.057  \\
    & Task Data & 0.073 & -0.057 &\cellcolor{apricot} 0.115* & \cellcolor{gray} \\
    & Lang Sim. & \cellcolor{raspberry} 0.475** & \cellcolor{raspberry} 0.518** & \cellcolor{raspberry} 0.492** & 0.076  \\
    \hline 
    \multirow{4}{*}{Forgetting} & Src. Pretraining Data & \cellcolor{raspberry} -0.208** & \cellcolor{apricot} -0.123* & 0.000 & \cellcolor{apricot}  0.137* \\
    & Trg. Pretraining Data & 0.042 & 0.015 & \cellcolor{apricot} 0.122* & -0.079 \\
    & Task Data & 0.009 & -0.004 & 0.078 & \cellcolor{gray} \\
    & Lang Sim. & \cellcolor{apricot} 0.165* & \cellcolor{apricot} 0.186* & -0.025 & \cellcolor{apricot} 0.164*  \\
    \toprule
    \end{tabular}
    \caption{Correlation study of different factors against measures of cross-lingual transfer. \hlone{* p < 0.05}, \hltwo{** p < 0.001}}
    \label{tab:x-lang-corr}
\end{table*}

\section{What Factors Affect Multilingual Learning?}
\label{app:correlations}
This section presents extended results analyzing the correlations between different factors and the in-language and cross-lingual learning exhibited by XLM-R$_{replica}$.

\subsection{In-language Correlation Study}
\label{app:in-lang-corr}
We consider whether the following factors correlate with various measures of model learning (Table \ref{tab:in-lang-corr}): \textit{pretraining data}, the amount of text in the CC100 corpus for each language; \textit{task data}, the amount of in-task data used to train each probe; and \textit{language similarity} to English, which is the highest-resource language in the pretraining data. We use the syntactic distances calculated in \citet{malaviya17emnlp} as our measure of language similarity; these scores are smaller for more similar language pairs. 

\paragraph{Overall Performance} The amount of pretraining data and in-task training data are strongly correlated with overall task performance for most of the considered tasks; this corroborates similar results from \citet{wu2020all}. Language similarity with English is also correlated with better in-task performance on all tasks except for dependency arc prediction, suggesting that some form of cross-lingual signal supports in-language performance for linguistically similar languages.

\paragraph{Learning Progress Measures} We also consider (1) the step at which XLM-R$_{replica}$ achieves 95\% of its best performance for each language and task, which measures how quickly the model obtains a majority of the tested linguistic information, and (2) how much the model \textit{forgets} from the best performance for each language by the final training checkpoint. We find that language similarity to English is strongly correlated with how quickly XLM-R$_{replica}$ converges on BPC and dependency arc classification. This suggests that cross-lingual signal helps the model more quickly learn lower-resource languages on these tasks, in addition to improving overall model performance. However, we observe no strong trends as to what factors affect forgetting across tasks.

%Is speed of acquisition correlated with overall performance? No

\subsection{Cross-lingual Correlation Study}
\label{app:cross-lang-corr}
Table \ref{tab:x-lang-corr} presents a correlation study of different measures for cross-lingual transfer in XLM-R$_{replica}$. We consider the effect of source and target pretraining data quantity, the amount of in-task training data (in the source language), and the similarity between the source and target language on the following transfer measures: overall task performance, asymmetry in transfer (the difference in model performance on $l_1 \rightarrow l_2$ compared to $l_2 \rightarrow l_1$), the step at which the model achieves 95\% or more of overall performance on the language pair, and forgetting -- the (relative) degradation of overall performance in the final model checkpoint. 

\paragraph{Correlations of Transfer with Language Factors} For overall cross-lingual performance, we observe that language similarity is highly correlated with task performance for all tasks and is similarly correlated with speed of acquisition (the step to 95\% of overall performance) for three of the four considered tasks. This is in line with prior work that has also identified language similarity as a strong indicator of cross-lingual performance \cite{pires2019multilingual}.
However, all considered factors are less correlated with the other measures of knowledge acquisition, such as the asymmetry of transfer and the forgetting of cross-lingual knowledge; this suggests that there could be other factors that explain these phenomena.

\paragraph{Interactions Between Learning Measures} We also consider the correlations between the different measures of model performance on cross-lingual transfer. For example, overall transfer performance is strongly correlated (p<<0.001) with earlier acquisition (step to 95\% of overall performance) for all syntactic tasks: {$\rho=-0.50$} for both POS tagging and dependency arc prediction and $-0.55$ for arc classification. To a lesser extent, overall transfer performance and model forgetting are negatively correlated, ranging from $\rho=-0.13$ to $-0.42$ across considered tasks. This indicates that XLM-R$_{replica}$ forgets less of the learned cross-lingual signal for better-performing language pairs, at the expense of already less successful ones. 

\begin{figure}
    \centering
    \includegraphics[width=\linewidth]{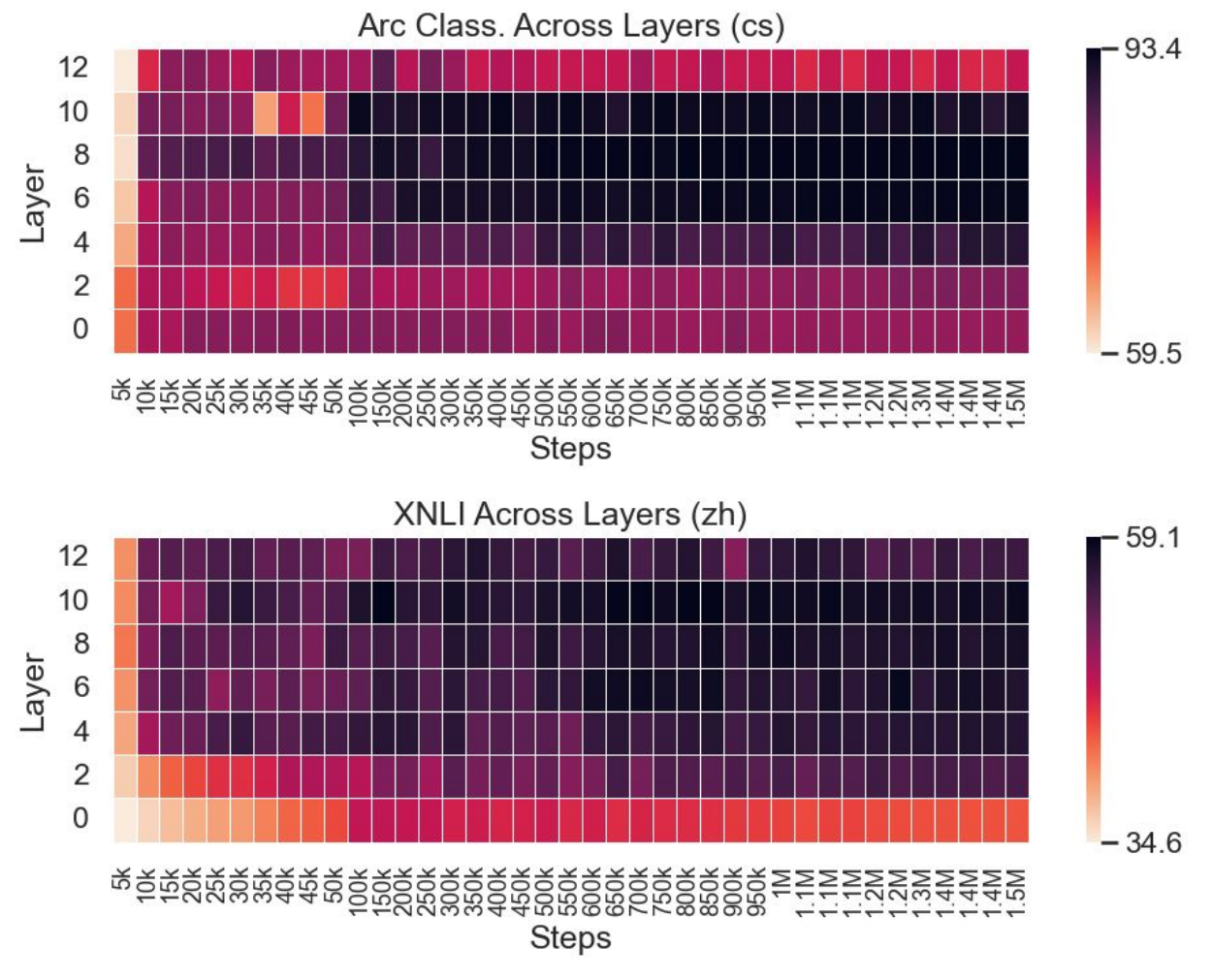}
    \caption{Layer-wise performance heatmaps for Czech arc classification and Chinese XNLI.}
    \label{fig:app-in-lang-layer}
\end{figure}

\begin{figure}
    \centering
    \includegraphics[width=\linewidth]{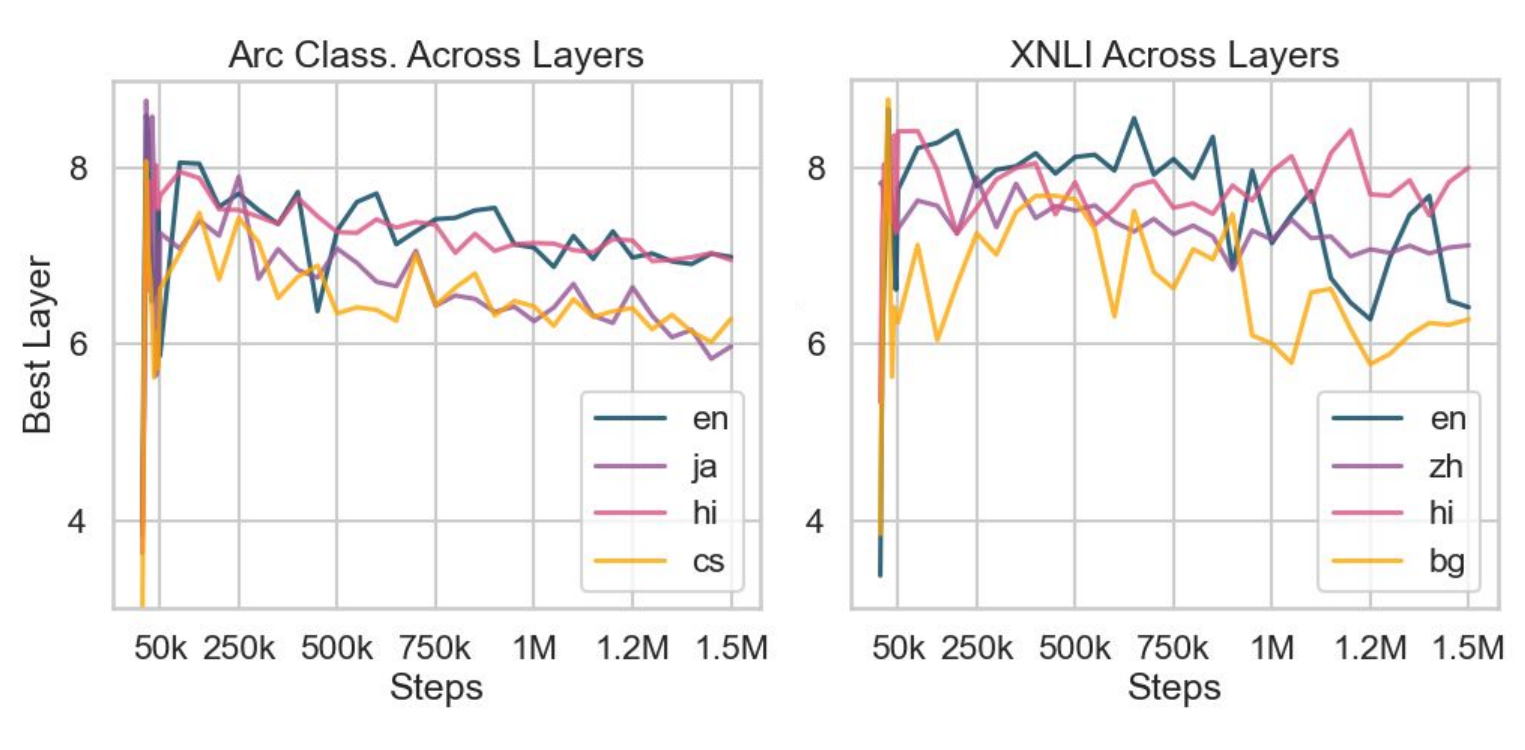}
    \caption{The expected best layer for in-language dependency arc classification and XNLI over time on XLM-R$_{replica}$.}
    \label{fig:app-exp-layer}
\end{figure}

\begin{figure}
    \centering
    \includegraphics[width=\linewidth]{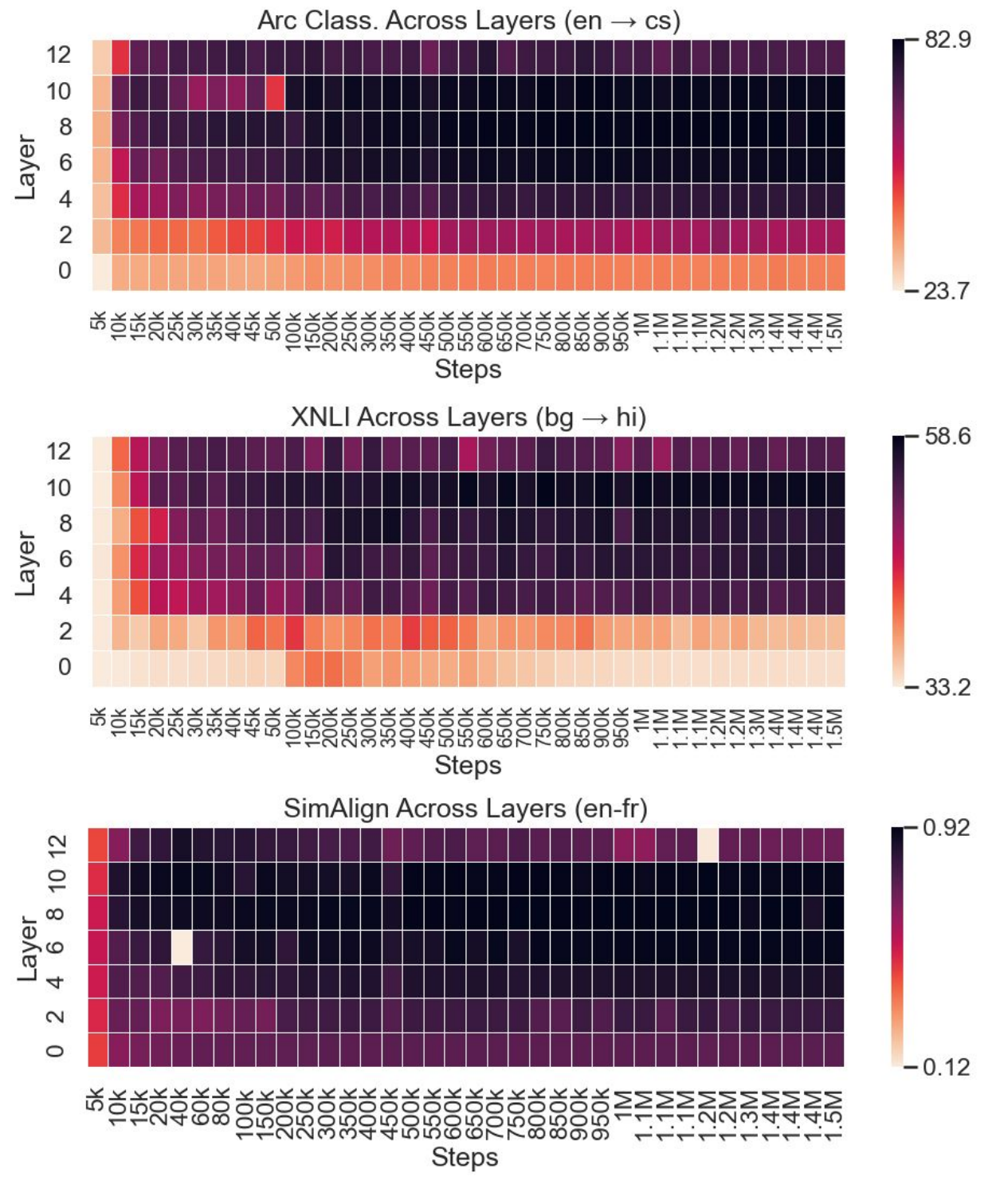}
    \caption{Additional heatmaps of cross-lingual transfer at different layers and timesteps of XLM-R$_{replica}$.}
    \label{fig:app-xl-layer}
\end{figure}

\begin{figure}
    \centering
    \includegraphics[width=\linewidth]{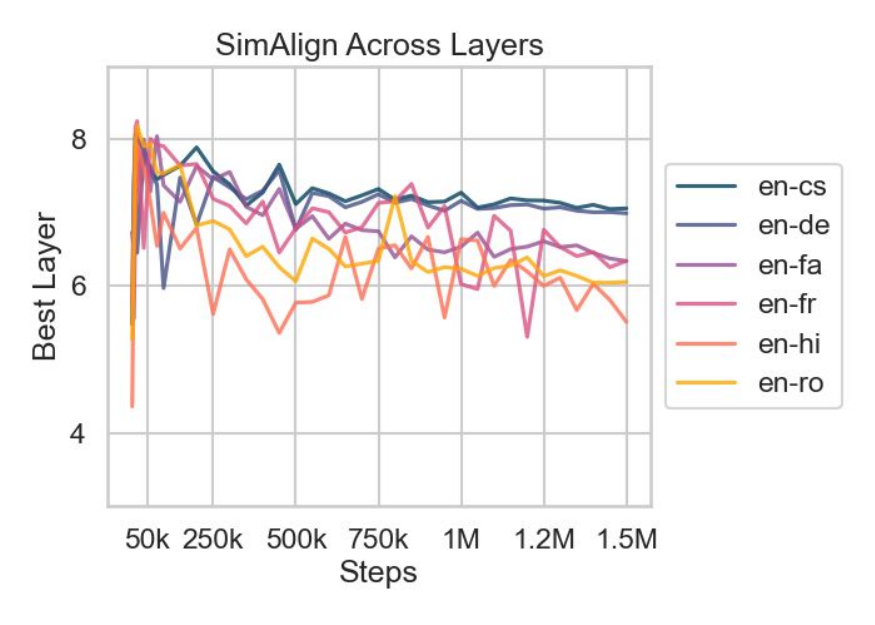}
    \caption{Change in the expected best layer for word alignment via SimAlign over time in XLM-R$_{replica}$}
    \label{fig:app-simalign-exp-layer}
\end{figure}

\section{Expanded Layer-wise Analysis}
\label{app:layerwise}
This section expands on the layer-wise analysis of XLM-R$_{replica}$ presented in \S \ref{sec:layerwise}. Figure \ref{fig:app-in-lang-layer} gives additional layer-wise heatmaps over time. Figure \ref{fig:app-exp-layer} shows the expected layer (i.e., average layer weighted by relative performance) of XLM-R$_{replica}$ at different time steps. The expected layer decreases over time: by 1.79, 1.61, 1.08, and 2.79 for CS, EN, HI, and JA respectively on dependency arc classification; and by 2.49, 2.25, 0.43, and 0.77 for BG, EN, HI, and ZH respectively on XNLI.

We also provide additional examples of layer-wise cross-lingual transfer in Figure \ref{fig:app-xl-layer}; we find that for cross-lingual transfer,  the best internal layer outperforms the best final layer state on average by 7.67 on arc classification transfer, 3.39 on XNLI, and 14.2 F1 on Simalign. Figure \ref{fig:app-simalign-exp-layer} shows the change in the expected best layer over time for SimAlign.

\section{Additional Across Time Analyses}
\label{app:more-analysis}
\begin{figure}
    \centering
    \includegraphics[width=\linewidth]{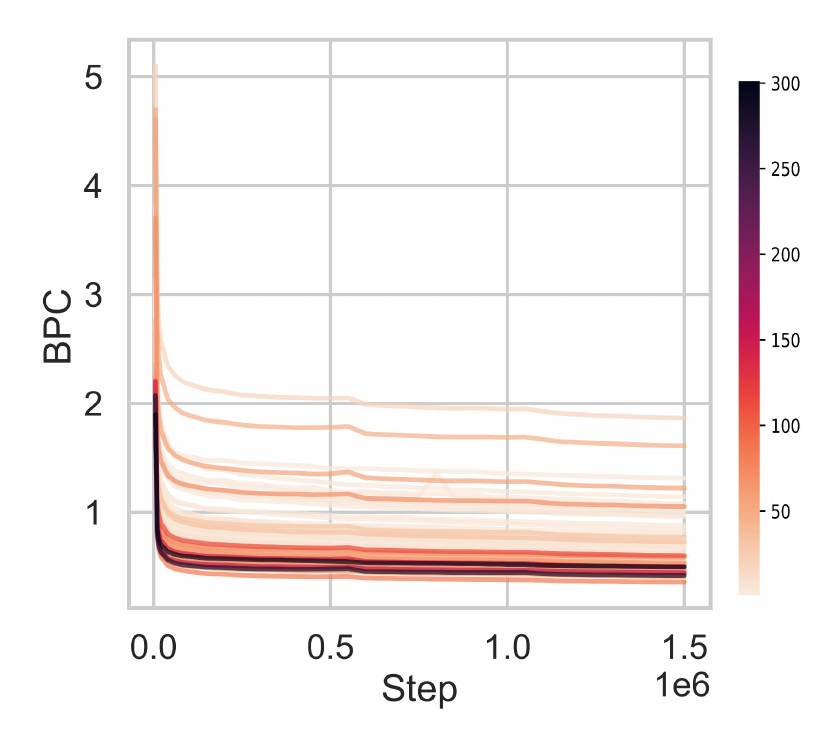}
    \caption{Learning Curves for BPC in each training language. Lines are colored by the amount of pretraining data available for that language.}
    \label{fig:bpc_across_time}
\end{figure}

\begin{figure}[b!]
    \centering
    \includegraphics[width=0.75\linewidth]{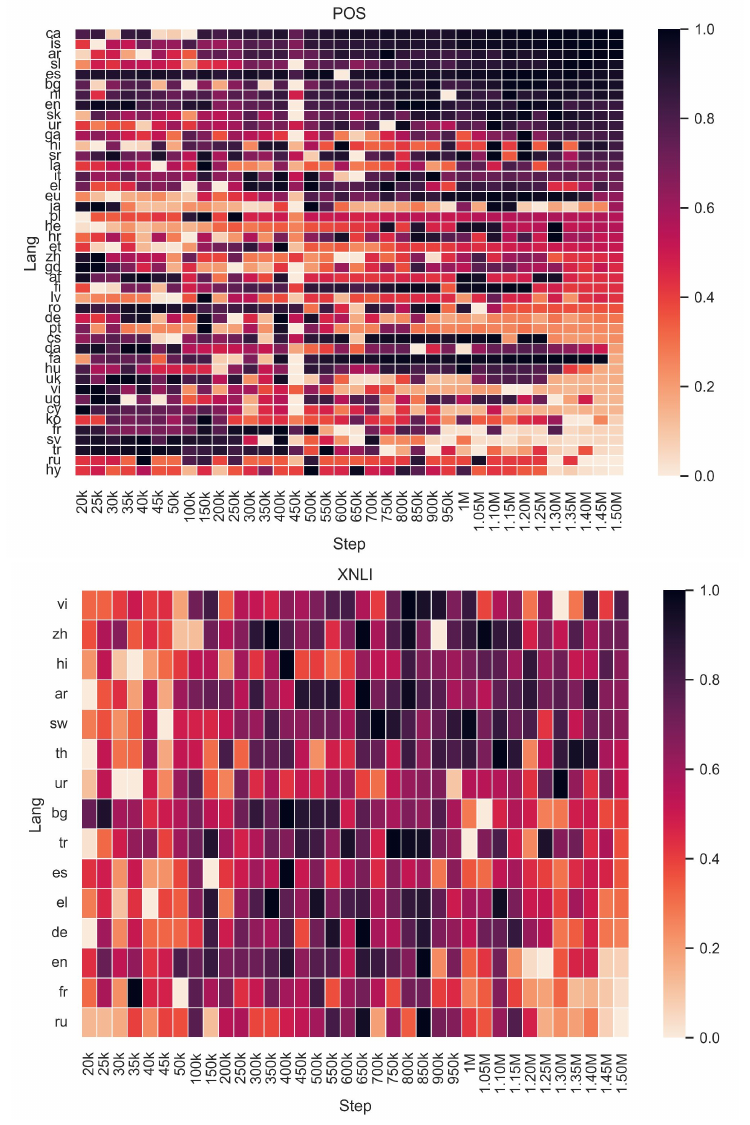}
    \caption{Heatmap of relative performance over time for different languages for POS tagging and XNLI. Languages are ordered by the amount of performance degradation at the final checkpoint.}
    \label{fig:app_progress_heatmap}
\end{figure}

\begin{figure*}
    \centering
    \includegraphics[width=\linewidth]{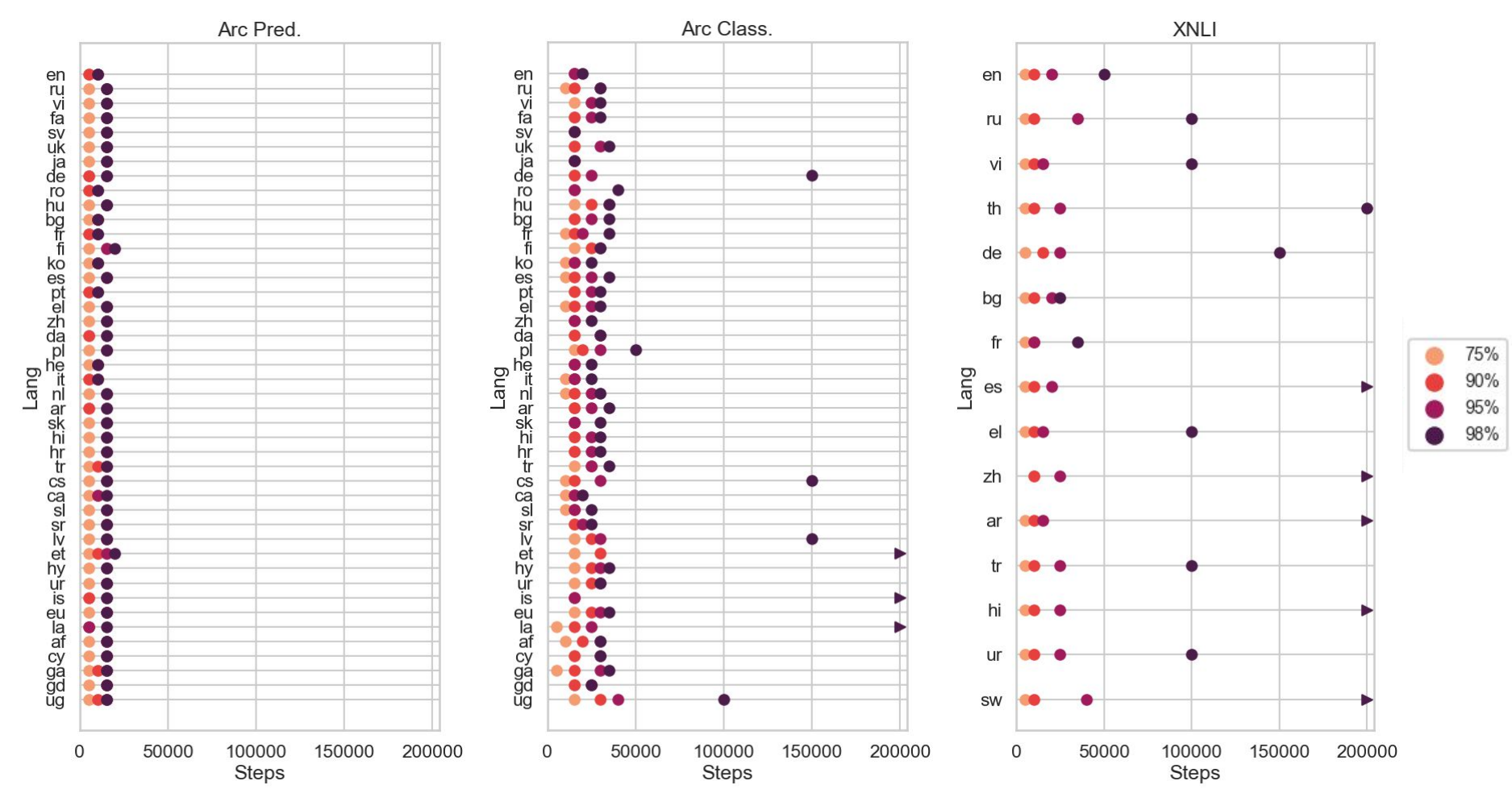}
    \caption{Learning Progress of XLM-R$_{replica}$ across training, up to 200k training steps. Each point represents the step at which the model achieves x\% of the best overall performance of the model on that task.}
    \label{fig:app_progress_scatter}
\end{figure*}

\begin{figure*}[t!]
    \centering
    \includegraphics[width=0.75\linewidth]{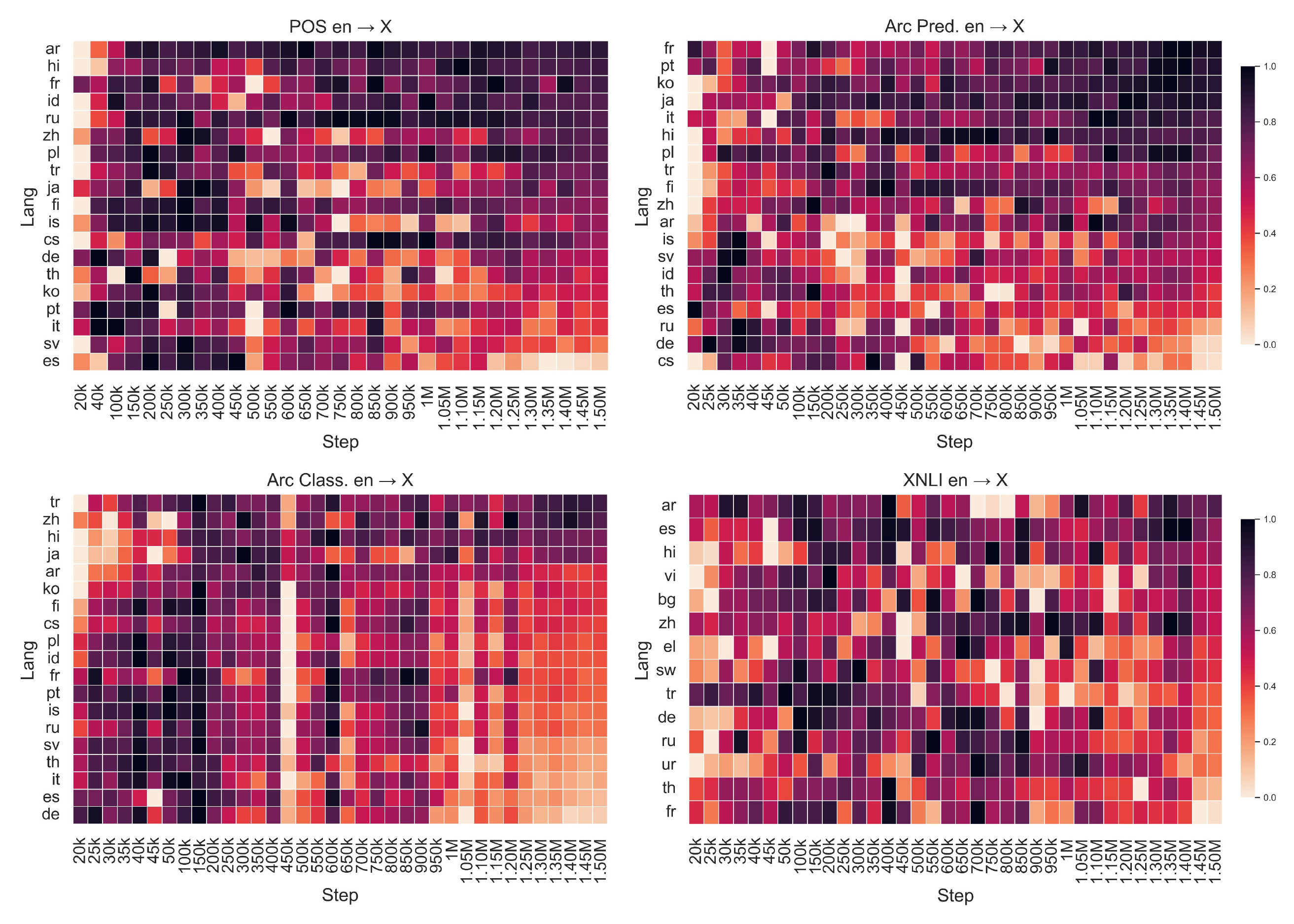}
    \caption{Heatmap of relative performance over time for cross-lingual transfer with English as the source language. Languages are ordered by the amount of performance degradation at the final checkpoint.}
    \label{fig:app_en_transfer_heatmap}
\end{figure*}

This section includes additional results from our analysis of knowledge acquisition during multilingual pretraining: 
\begin{itemize}
    \item Figure \ref{fig:bpc_across_time} presents BPC learning curves for each language in the CC100 training data.
    \item Figure \ref{fig:app_progress_scatter} covers the learning progress of XLM-R$_{replica}$ on dependency arc prediction, arc classification, and XNLI, expanding on the results in \S \ref{sec:when-in-lang}.
    \item Figure \ref{fig:app_progress_heatmap} gives the relative performance for in-language POS and XNLI across training checkpoints discussed in \S \ref{sec:when-in-lang}.
    %\item Figure \ref{fig:app_learning_curves} presents learning curves for a subset of languages on the four in-language tasks; these curves correspond to the summaries for those languages in Figures \ref{fig:app_progress_scatter} and \ref{fig:app_progress_heatmap}.
    \item Figure \ref{fig:app_en_transfer_heatmap} presents more detailed results for relative performance over time when transferring out of English. This expands on the summary figures discussed in \S \ref{sec:when-transfer}.
\end{itemize}

%\begin{figure*}
%    \centering
%    \includegraphics[width=\linewidth]{app-figures/app-learning-curves.pdf}
%    \caption{Learning curves over time for a subset of languages for each of the four in-language linguistic information evaluations.}
%    \label{fig:app_learning_curves}
%\end{figure*}

\end{document}